\documentclass[letterpaper, 10 pt, conference]{ieeeconf}  
\IEEEoverridecommandlockouts
\usepackage{amsmath,amssymb,amsfonts}
\usepackage{balance}
\usepackage{soul}
\usepackage[caption=false]{subfig}
\usepackage{graphicx}
\usepackage{textcomp}
\usepackage{multirow} 
\usepackage{xcolor}
\usepackage{float} 
\usepackage{tabularray}
\usepackage{array}
\usepackage{booktabs}
\usepackage{verbatim} 
\usepackage{makecell}
\usepackage{url}
\usepackage{algorithm}
\usepackage{algpseudocode}
\makeatletter

\newcommand{\Rmnum}[1]{\expandafter\@slowromancap\romannumeral #1@}

\makeatother

\def\BibTeX{{\rm B\kern-.05em{\sc i\kern-.025em b}\kern-.08em
    T\kern-.1667em\lower.7ex\hbox{E}\kern-.125emX}}

\title{\LARGE \bf Ultrasound-Guided Robotic Blood Drawing and In Vivo Studies on Submillimetre Vessels of Rats}
\author{Shuaiqi Jing$^{1, 2}$, Tianliang Yao$^{1}$, Ke Zhang$^{2}$, Di Wu$^{3}$, Qiulin Wang$^{2}$, Zixi Chen$^{4}$, Ke Chen$^{1, 2}$, Peng Qi$^{1, 5, *}$
\thanks{This work has been submitted to the IEEE for possible publication. Copyright may be transferred without notice, after which this version may no longer be accessible.}
\thanks{This work is supported by the National Key Research and Development Program of China under Grant No. 2023YFB4705200, the National Natural Science Foundation of China under Grant No. 62273257, and the Open Project Fund of State Key Laboratory of Cardiovascular Diseases No.2024SKL-TJ002.\emph{(*Corresponding Author: Peng Qi)}.}
\thanks{$^{1}$Department of Control Science and Engineering, College of Electronics and Information Engineering, and Shanghai Institute of Intelligent Science and Technology, Tongji University, No. 1239, Siping Road, Shanghai 200092, China;}%
\thanks{$^{2}$Chengdu Aixam Medical Technology Co., Ltd., Chengdu 610200, Sichuan Province, China;}%
\thanks{$^{3}$Katholieke Universiteit Leuven, Leuven 3000, Belgium;}%
\thanks{$^{4}$The BioRobotics Institute, Scuola Superiore Sant'Anna, Pontedera 56025, Italy;}%
\thanks{$^{5}$State Key Laboratory of Cardiovascular Diseases and Medical Innovation Center, Shanghai East Hospital, School of Medicine, Tongji University 200092, Shanghai, China.}%
}

\begin{document}

\maketitle
\pagestyle{empty}  
\thispagestyle{empty} 

\begin{abstract}
Billions of vascular access procedures are performed annually worldwide, serving as a crucial first step in various clinical diagnostic and therapeutic procedures. For pediatric or elderly individuals, whose vessels are small in size (typically 2 to 3 mm in diameter for adults and $<$1 mm in children), vascular access can be highly challenging. This study presents an image-guided robotic system aimed at enhancing the accuracy of difficult vascular access procedures. The system integrates a 6-DoF (Degrees of Freedom) robotic arm with a 3-DoF end-effector, ensuring precise navigation and needle insertion. Multi-modal imaging and sensing technologies have been utilized to endow the medical robot with precision and safety, while ultrasound (US) imaging guidance is specifically evaluated in this study. To evaluate in vivo vascular access in submillimeter vessels, we conducted ultrasound-guided robotic blood drawing on the tail veins (with a diameter of 0.7 ± 0.2 mm) of 40 rats. The results demonstrate that the system achieved a first-attempt success rate of 95\%. The high first-attempt success rate in intravenous vascular access, even with small blood vessels, demonstrates the system’s effectiveness in performing these procedures. This capability reduces the risk of failed attempts, minimizes patient discomfort, and enhances clinical efficiency.
\end{abstract}


\section{Introduction}
Vascular access is a crucial first step in numerous medical practices, providing a fundamental prerequisite for the subsequent execution of diagnostic and therapeutic interventions \cite{galena1992complications}. It is widely utilized across various clinical contexts for blood sampling, intravenous therapy, and medication administration. 
Nevertheless, acquiring vascular access can be extremely challenging in difficult conditions, particularly when dealing with small and fragile veins, typically less than 1mm in diameter. This scenario is commonly encountered in pediatric patients \cite{wong2023effects}. A study involving 150 pediatric patients revealed that the first-attempt success rate for vascular access was below 50\% in children aged 0-12 years \cite{hess2010biomedical}.
These unsuccessful punctures could result in vessel damage and subsequent secondary effects, such as increased risks of further vein damage and infection.

To mitigate these risks and enhance the success rate of vascular access, various imaging modalities, including near-infrared (NIR) light \cite{cuper2011visualizing}, ultrasound (US) \cite{maecken2007ultrasound, brusciano2024advantages}, and visible light \cite{katsogridakis2008veinlite}, have been integrated to assist clinicians in precisely locating veins. NIR imaging technology is capable of producing a more distinct two-dimensional outline of blood vessels, whereas a more costly US device is able to provide additional depth information. However, clinicians are still required to personally interpret these images to obtain information crucial for guiding puncture procedures. With the recent advancements in deep learning (DL) and computer vision (CV) technologies, algorithms can now be utilized for image processing, significantly reducing the cognitive load on clinicians. Nevertheless, irrespective of whether image interpretation is manual or AI-based, the clinician still has to manually perform the puncture. Additionally, as mentioned above, some imaging systems only offer two-dimensional representations of blood vessels, lacking depth perception and potentially insufficient resolution to detect subtle vein movements, which could potentially complicate the puncture process. Furthermore, it has been suggested that the greatest challenge in venous cannulation lies not merely in the initial identification of the vein, but rather in the delicate task of accurately inserting the needle or cannula \cite{balter2015system}.


\begin{figure}[t]
    \centering
    \includegraphics[width=0.96 \linewidth]{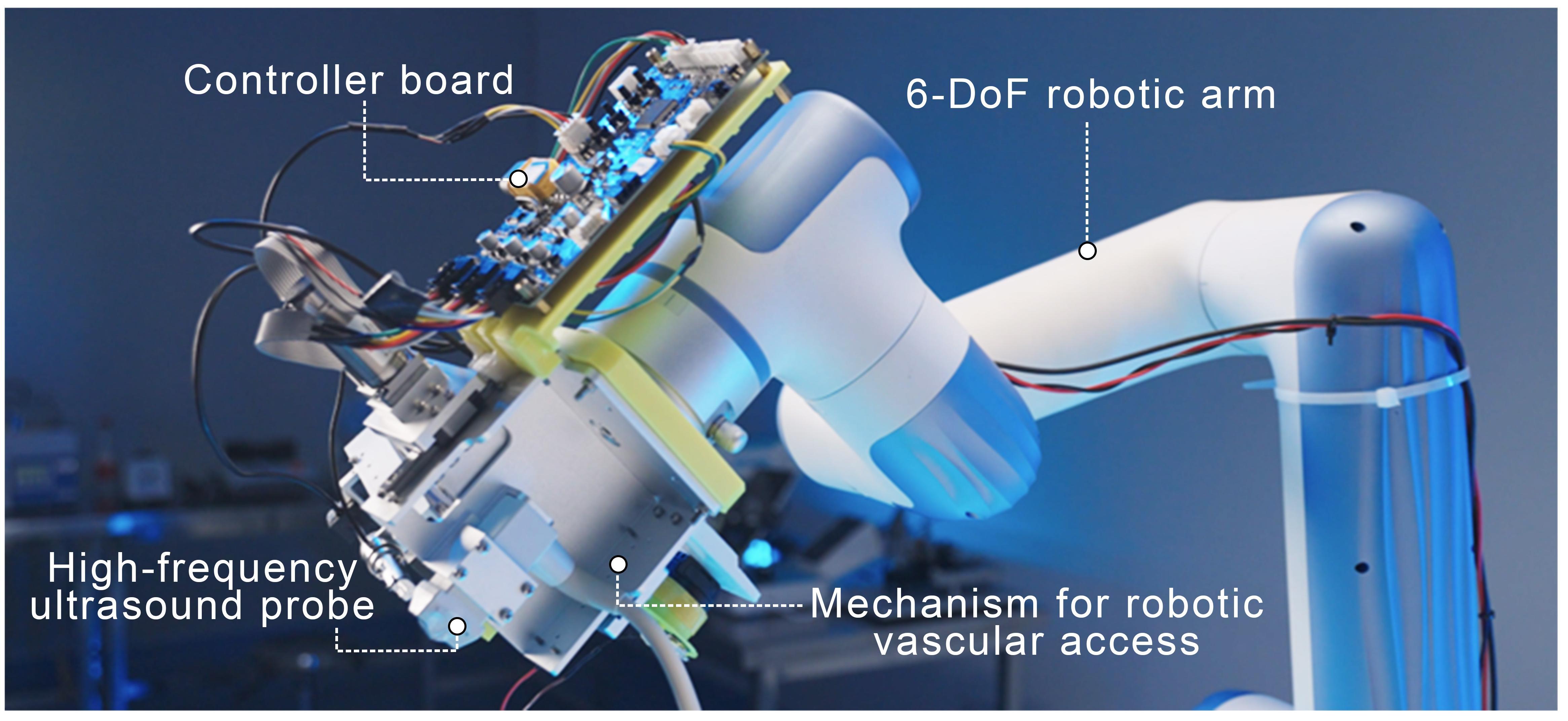}
    \caption{Prototype of the proposed Ultrasound-guided vascular access robotic system. The system comprises a 6-DoF robotic arm integrated with a specialized mechanism for robotic vascular access, enabling precise needle manipulation under ultrasound guidance.}
    \label{fig:system}
\end{figure}


Robotics has proven to offer advantages in medical procedures, achieving a level of precision and consistency that is comparable to, and in some cases exceeds, that of clinicians \cite{Guangzhong2017}. The enhanced precision of robotic systems minimizes the risk of errors and complications, potentially leading to higher success rates and better patient outcomes. These benefits render the integration of medical robotics into vascular access procedures particularly promising, as it holds the potential to significantly improve both the efficiency and reliability of this common medical practice. Several pioneering and notable research works have been reported, with commercial product developments also advancing concurrently \cite{vitestro2024}. For example, Balter and Chen \textit{et. al.} devised a 7-degree-of-freedom (DOF) robotic system for executing venipuncture \cite{balter2015system}. This system comprises a 3-DOF gantry aimed at imaging the patient's peripheral forearm veins, coupled with a compact 4-DOF serial arm that guides the cannula into the targeted vein under closed-loop control. The robot incorporates both near-infrared and ultrasound imaging capabilities, along with an image processing unit. Nevertheless, the system's validation was confined to a phantom arm, with cannulated vessels measuring 3.2 mm and 1.8 mm in diameter, thereby leaving its efficacy in actual clinical settings untested. Subsequently, the system underwent further development and in vivo validation on rats \cite{chen2020deep}, yielding a notable enhancement in first-attempt access rates. The robotic cannulation achieved an 87.1\% success rate on the initial attempt, compared to a 58.3\% success rate in the control group that relied on manual insertion. Besides, another 7-DOF trocar puncture robot was introduced to facilitate blood collection \cite{yang2024trocar}. This robot boasts a position and orientation adjustment mechanism, along with an end-effector for precise needle manipulation. The development encompassed both inverse kinematics and path planning algorithms for the robot. However, the system's validation was limited to a phantom model with a simulated vein of 4 mm in diameter, hindering its applicability in pediatric cases. Furthermore, the robot system lacks integration with any imaging modality, which further constrains its clinical validation and practicality. Additionally, these robotic systems still necessitate expertise from both medical and robotics professionals for accurate positioning and calibration.
In response to the challenges mentioned above, coupled with the epidemic that has spurred the demand for more contactless and unmanned medical services \cite{Guangzhongcovid}, intensive preliminary studies on robotic vascular access have been conducted in recent years, encompassing the authors' prior research work in this field \cite{chen2021semi, huang2021autonomous, ji2022automated}. This paper introduces a newly upgraded robotic setup, which is potentially undergoing commercialization, as illustrated in Fig. \ref{fig:system}. This system employs multi-modal imaging and sensing technologies, integrated with a readily available robotic arm, to facilitate vessel access with minimal human intervention, thereby mitigating reliance on operator skill. Subsequent sections commence with a comprehensive overview of the system design, encompassing imaging/sensing modules, the needle insertion mechanism, and the control architecture. Notably, we assess the performance of the system in US-guided procedures for difficult vascular access on both phantoms and animals, underscoring its effectiveness and potential for clinical application. This evaluation highlights the system's potential to offer a safer and more effective solution, particularly in pediatric settings.



The main contributions of this paper are:
\begin{itemize}
\item A newly designed, fully functional robotic platform to perform vascular access.
\item A high first-attempt success rate with precise \textit{in vivo} blood draws from submillimeter vessels in rats.
\end{itemize}


\section{System Design and Integration}
\subsection{System Overview}
\begin{figure}[t]
    \centering
    \includegraphics[width=0.92 \linewidth]{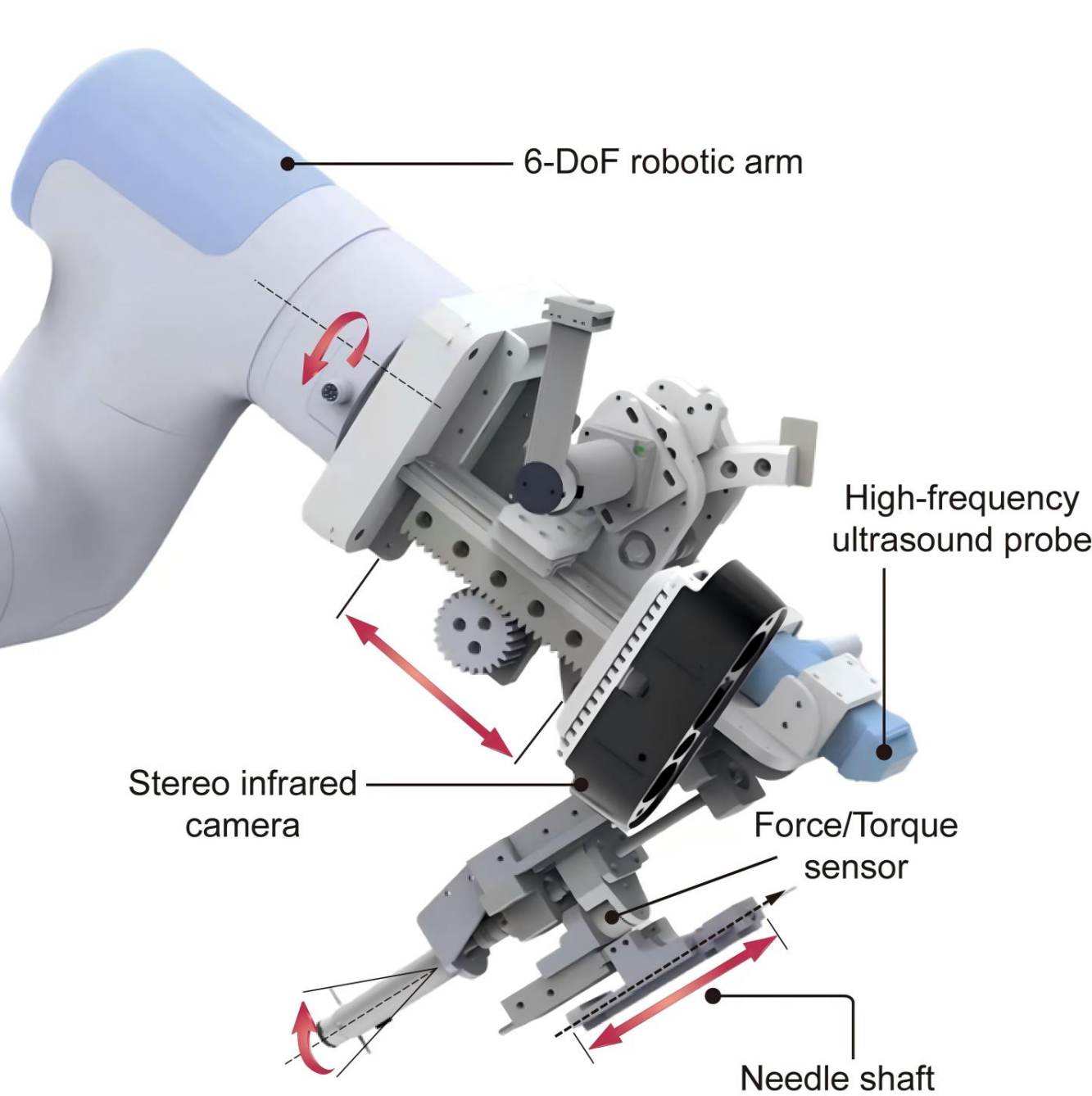}
    \caption{Computer-aided design renders to illustrate the distal manipulator details.}
    \label{fig:RoboticSystem}
\end{figure}

As illustrated in Fig. \ref{fig:RoboticSystem}, the proposed Robotic Vascular Access (RVA) system has been designed to maneuver and guide the needle precisely into the target vessel. The system has several components, including: a 6-Degree-of-Freedom (DoF) off-the-shelf robotic arm for accurate spatial positioning, a high-precision motion control system, a force/torque sensor for delicate tissue interactions, a stereo near-infrared (NIR) camera for real-time tracking and registration, and a high-frequency ultrasound (US) transducer for enhanced tissue imaging and depth guidance. The rationale for selecting an existing, commercial robotic arm is twofold. Firstly, it possesses a broader range of motion and positioning capabilities compared to a custom-made platform, enabling more versatile and expansive operations. Secondly, a commercial robotic arm is inherently capable of achieving higher and more consistent positioning accuracy, thereby ensuring precision and reliability in conducting the tasks.

\subsection{Ultrasound Imaging and Vessel Detection}

The RVA system enhances its NIR imaging module with advanced ultrasonic imaging to precisely determine the location, size, and depth of target veins and surrounding anatomy, significantly improving the accuracy and safety of vascular access procedures. The ultrasound subsystem employs short-axis and long-axis guidance techniques \cite{amel2023toward}. Binocular NIR imaging captures rat tail vein images, extracting the vein’s approximate position from NIR images to directly guide the ultrasound probe’s alignment with the target, thereby facilitating high-fidelity ultrasound imaging. In this paper, the effectiveness of the application of ultrasound imaging will be the main focus.


In the short-axis view, the ultrasound probe is oriented perpendicular to the skin surface above the target vessel, generating a cross-sectional image. This approach is particularly beneficial for distinguishing between veins and arteries and allows for real-time monitoring of the needle's entry into the vessel. Conversely, the long-axis view aligns the ultrasound probe parallel to the vessel, presenting a longitudinal image that simplifies continuous tracking of the needle along the vessel's trajectory, thus minimizing the risk of needle deviation during insertion.

Imaging superficial and small-caliber veins, particularly in pediatric patients like infants and young children, poses unique challenges stemming from their diminutive size and proximity to the skin surface. To tackle these challenges, the RVA system incorporates clinical guidelines and expert knowledge to refine the ultrasound imaging approach. It employs a high-frequency 12.4 MHz ultrasound transducer, mounted on a robotic arm, which is utilized to attain an optimal short-axis imaging position precisely above the target vessel, guaranteeing a clear and stable cross-sectional view.

Given the distinctive characteristics of superficial fine vessels in pediatric patients, the ultrasound imaging parameters are meticulously calibrated. The system utilizes a high gain, a reduced dynamic range, and a customized grayscale map to amplify the visibility of small vessels, simplifying their detection and subsequent path planning. The technical specifications incorporate a gain setting of 80 dB, a depth setting of 1.6 cm, a narrow dynamic range of 80 dB, and an operating frequency of 14.2 MHz. Image enhancement is set to level 3, accompanied by a grayscale map adjustment of 14 and a frame correlation of 2. These precise parameter adjustments are pivotal in augmenting the clarity of fine vascular structures, ultimately enabling more precise and safer puncture guidance.

\subsection{The 9-DoF Robotic System}

The widespread adoption of 6-DoF collaborative robotic arms is driven by their versatility, extensive operational range, and high precision, rendering them indispensable in diverse clinical applications, including laparoscopic surgery and orthopedic procedures \cite{kuo2012kinematic}, as previously explained to some extent. To cater to the specific requirements of intravenous vascular access, a 6-DoF robotic arm is integrated with a 3-DoF end-effector specifically designed for vascular access, thereby constituting a system with a total of 9-DoF.





As illustrated in Fig. \ref{fig:RoboticSystem}, the distal manipulator can be rotated via the last joint; a gear transmission mechanism is used to achieve its forward-and-backward movement, which is utilized for linearly moving the ultrasound probe. The needle is fixed in a designated position, and its pitch angle can be adjusted through the mechanical design presented here. Finally, the needle shaft can be controlled for insertion and retraction, while a force/torque sensor attached to its end enables detection.

The 3-DoF end-effector, equipped with an ultrasound probe and a needle, facilitates precise control over positioning and orientation, essential for accurate vascular imaging and needle insertion. The integration of high positional accuracy with real-time ultrasound feedback ensures minimal deviation during needle placement, thereby mitigating the risk of complications.

\subsection{Needle Insertion Procedure}
The RVA system operates through a meticulously crafted workflow, guaranteeing precise and safe needle insertion. This workflow, outlined in Algorithm 1, comprises five pivotal steps: pre-operative calibration, initial positioning, target identification and alignment, path planning and insertion, and post-procedure reset. Specifically, each step can be elaborated upon as follows:

The procedure commences with pre-operative calibration to ascertain the needle's precise positioning. The system employs NIR scanning of the patient's arm, positions the robotic arm, and fine-tunes the ultrasound settings for optimal image clarity. Upon identifying the target vein within the ultrasound image, the system aligns the needle using inverse kinematics. Subsequently, the needle is inserted along a pre-planned trajectory, with continuous monitoring of insertion force and tissue deformation to uphold safety and precision. Should any complications arise during insertion, such as excessive force, the system promptly adjusts the trajectory or retries the insertion. Upon completion of the procedure, the needle is safely withdrawn, and the robotic arm returns to its designated home position, thereby concluding the entire process. Each step is indispensable in attaining the accuracy and reliability demanded in clinical settings, minimizing the risk of complications, and safeguarding patient safety.

\begin{table}[ht]
\caption{Nomenclature employed in the intravenous vascular access system algorithm}
\label{tab:nomenclature}
\centering
\begin{tabular}{cc}
\hline
\textbf{Symbol} & \textbf{Description} \\
\hline
$T_{cal}, T_{expected}$ & \makecell[c]{Calibration and expected transformation matrices} \\
$\epsilon_{cal}$ & Calibration error threshold \\
$\mathbf{x}, \mathbf{q}$ & End-effector position and joint angle vector \\
$f(\cdot), f^{-1}(\cdot)$ & \makecell[c]{Forward and inverse kinematics functions} \\
$Q, Q_{threshold}$ & Image quality score and threshold \\
$\mathbf{p}_{target}, \mathbf{p}_{current}$ & Target and current positions \\
$\epsilon_{align}$ & Alignment error threshold \\
$\mathbf{p}(t), \mathbf{v}$ & \makecell[c]{Position trajectory at time $t$ and insertion velocity} \\
$\mathbf{F}(t), K$ & Insertion force at time $t$ and tissue stiffness matrix \\
$\mathbf{u}(t), \epsilon_{deform}$ & \makecell[c]{Tissue deformation at time $t$ and deformation \\ threshold} \\
$F_{threshold}$ & Force threshold for insertion \\
$\mathbf{q}_{home}$ & Robot home position \\
\hline
\end{tabular}
\end{table}

\begin{algorithm}
\caption{Operational Procedure for Robotic Vascular Access System}
\begin{algorithmic}[1]
\State \textbf{Pre-operative Calibration:}
\State Calibrate needle and compute transformation matrix $T_{cal}$
\State Abort if $|T_{cal} - T_{expected}| > \epsilon_{cal}$
\State \textbf{Initial Positioning:}
\State Identify a suitable vessel via NIR
\State Position robotic arm to the designated area using forward kinematics: $\mathbf{x} = f(\mathbf{q})$
\State Adjust ultrasound and evaluate image quality $Q$
\State Retry if $Q < Q_{threshold}$
\State \textbf{Target Identification and Alignment:}
\State Identify target position $\mathbf{p}_{target}$ in ultrasound image
\State Compute distance $d = |\mathbf{p}_{target} - \mathbf{p}_{current}|$
\State Solve inverse kinematics for alignment: $\mathbf{q}_{aligned} = f^{-1}(\mathbf{p}_{target})$
\State Adjust if $|\mathbf{q}_{aligned} - \mathbf{q}_{current}| > \epsilon_{align}$
\State \textbf{Path Planning and Insertion:}
\State Compute trajectory $\mathbf{p}(t) = \mathbf{p}_0 + t\mathbf{v}$
\State Solve inverse kinematics along trajectory: $\mathbf{q}(t) = f^{-1}(\mathbf{p}(t))$
\State Perform insertion while monitoring force $\mathbf{F}(t)$
\State Estimate tissue deformation: $\mathbf{u}(t) = K^{-1}\mathbf{F}(t)$
\State Adjust trajectory if $|\mathbf{u}(t)| > \epsilon_{deform}$
\State Retry insertion if $\max(\mathbf{F}(t)) > F_{threshold}$
\State \textbf{Post-procedure Reset:}
\State Retract needle along path $\mathbf{p}_{retract}(t) = \mathbf{p}(t_{final}) - t\mathbf{v}$
\State Return robotic arm to home position $\mathbf{q}_{home}$
\end{algorithmic}
\end{algorithm}

\begin{figure}[ht]
    \centering
    \includegraphics[width=0.94 \linewidth]{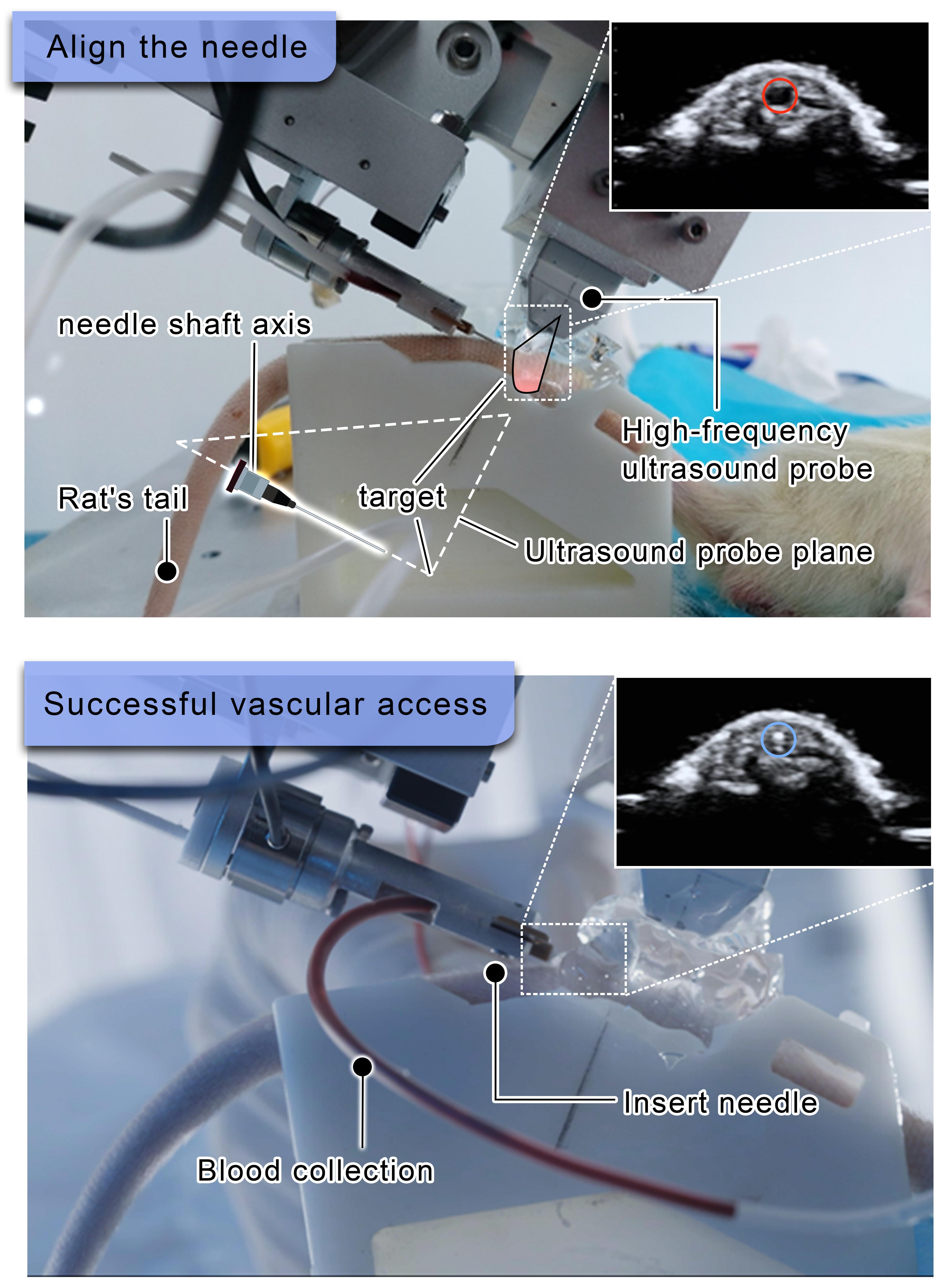}
    \caption{Experimental setup for the robotic and ultrasound-guided intravenous vascular access system: (a) \textit{In vivo} demonstration of venipuncture on a rat tail, illustrating the system's operational workflow during the procedure, and (b) Visualization of blood return in the catheter, confirming successful venipuncture, which is a critical metric for assessing system performance.}
    \label{fig:ExperimentalSetup}
\end{figure}

%
%




\section{Experiments}
\subsection{Experimental Setup}
The RVA system incorporates clinical guidelines and expert knowledge to enhance short-axis ultrasound imaging for superficial vascular examinations, leveraging the Stork$^\circledR$ Cloud Ultrasound Device. This device is mounted on a 6-DoF robotic arm equipped with a high-frequency 12.4 MHz probe, capable of achieving a resolution of 0.1 mm. The robotic arm is programmed to precisely position the probe vertically above the target vessel, ensuring clear and accurate cross-sectional imaging.

To validate the accuracy and reliability of the system in a clinical context, \textit{in vivo} experiments were conducted utilizing Sprague-Dawley (SD) rat tail veins, which mimic the characteristics of infant veins. The objective of these experiments was to confirm the system's efficacy in safely puncturing small, superficial veins. In these experiments, ultrasound imaging of the rat tail vein was optimized by tuning key parameters, using a liquid coupling agent for seamless acoustic coupling to minimize impedance-related reflection losses. Tissue heterogeneity and surface irregularities were managed through iterative probe repositioning and parameter recalibration. Settings included a gain of 80 dB for enhanced vascular visibility, an imaging depth of 1.6 cm to capture the vein’s full structure without signal loss, a dynamic range of 80 dB to sharpen vessel wall contrast, a frequency of 14.2 MHz for detailed superficial imaging, an image enhancement level of 3 for improved clarity, and a 14-level grayscale map for refined tissue differentiation. These adjustments enabled clear visualization of the vein’s wall layers and lumen, supporting subsequent procedures. As a preliminary validation step, a custom-made polymer phantom was employed, featuring simulated vascular channels with a diameter of 4 mm positioned 3 mm beneath the surface.


The robotic system, based on the Nova 2 collaborative arm (DOBOT$^\circledR$ Inc., China) with 6-DoF and a repeat positioning accuracy of ± 0.05 mm, integrates a custom-made 3-DoF end-effector designed for vascular access. This end-effector allows vertical translation, rotation, and precise needle insertion, achieving an absolute positioning accuracy of 0.1 mm and a repeat accuracy of 0.01 mm, ensuring high precision in intravenous vascular access tasks.

\begin{figure*}[ht]
\centering
\includegraphics[width=0.9 \linewidth]{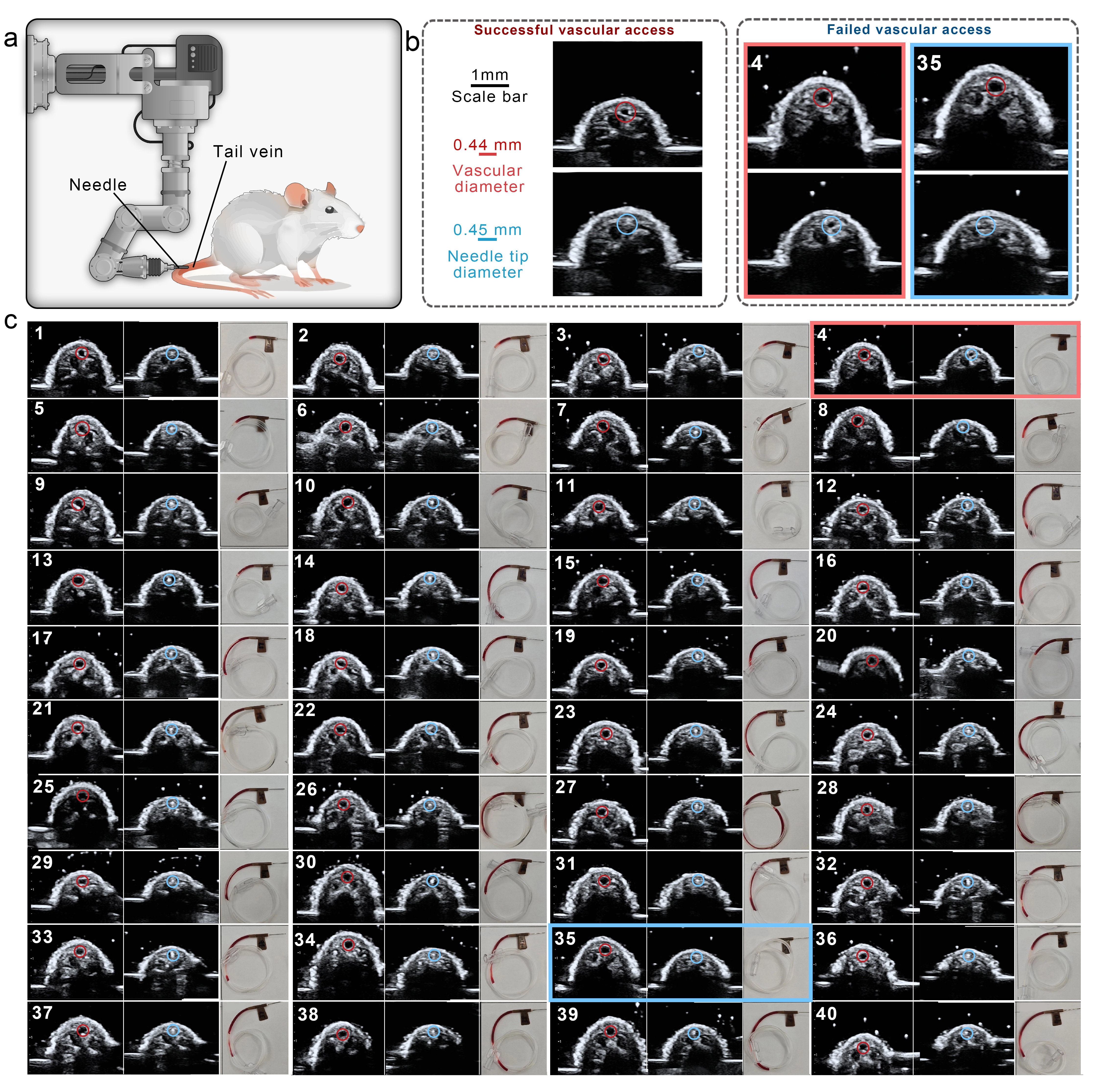}
\caption{Overview of the \textit{in vivo} robotic ultrasound-guided vascular access procedure in rat tail veins, showcasing the integration of robotic and ultrasound assistance for precise needle placement:
(a) Schematic illustration of the vascular access procedure, depicting the relative positioning of the needle and the rat tail.
(b) Representative ultrasound image from three of the 40 trials, emphasizing a transverse cross-section of the rat tail vein. The image distinctly identifies the vein's lumen and surrounding tissue, with the needle accurately positioned within the vein. The first column showcases a successful vascular access example: The upper ultrasound image depicts the pre-puncture scene, where the target vessel is visibly marked with a red circle. The lower image, post-puncture, reveals a bright spot in the previously targeted vessel, denoted by a blue circle, signifying the needle tip's location. The subsequent two columns illustrate failed puncture attempts: the fourth and 35th attempts. In both columns, the upper image clearly displays the target vessel, yet the corresponding lower image does not mirror the successful outcomes. Instead, the images indicate that the needle tip's bright spot either strays from the target vessel or remains undetected within it.
(c) The image grid encompasses all 40 trials, presenting pre-puncture and post-puncture ultrasound images, alongside the rightmost image in each row indicating the blood return status. This comprehensive result underscores the system's accuracy and reliability in achieving successful venous access.}
\label{fig:Ultrasound_Results}
\end{figure*}

\subsection{In-Vivo Validation}
Prior to the \textit{in vivo} validation, the system underwent initial validation through a phantom study utilizing a high-fidelity simulated human arm, custom-made from high-polymer material. The phantom was meticulously constructed using gellan gum to closely mimic the mechanical properties of human tissue \cite{morchi2021reusable}. The experimental protocol included performing a series of ten puncture attempts on each simulated vessel channel, with puncture sites randomly chosen to simulate clinical variability.
The system achieved a success rate of 100\% in all puncture attempts, unequivocally showcasing its exceptional accuracy and reliability within a controlled, simulated environment. The result lays a solid foundation for subsequent \textit{in vivo} animal studies.

The system underwent further validation through an \textit{in vivo} study employing Sprague-Dawley (SD) rats. SD rats were selected for this study owing to the anatomical similarity between the diameters of their tail veins and those of pediatric patients, rendering them an ideal model for evaluating potential pediatric applications of the system. The study encompassed 40 specific pathogen-free (SPF) grade SD rats, aged 10 weeks, with an average body weight of 280 ± 20 grams. Pre-experimental ultrasound imaging revealed an average tail vein diameter of 0.7 mm, confirming their suitability for this research. This study and its experimental procedures received approval from the Animal Care and Use Institutional Committee of NEW Biotechnology (Chengdu) Co., Ltd. All animal housing and experimental procedures were conducted in strict adherence to the institutional guidelines for the care and use of laboratory animals.

Each rat underwent two venipuncture attempts, targeting the proximal one-third to one-half segment of the tail vein with needle insertion proceeding from posterior to anterior. Successful venous access was confirmed by the visualization of blood return, followed by the injection of 100  $ \mu $L of physiological saline solution through the needle sheath.

The smallest vein successfully punctured had a short-axis diameter of 0.44 mm, while the average diameter of veins successfully punctured was recorded as 0.61 mm. Figure \ref{fig:Ultrasound_Results}(b) presents a representative ultrasound image of a rat tail vein in transverse cross-section, clearly outlining the lumen and its surrounding tissue.

Figure \ref{fig:Ultrasound_Results} also presents additional data highlighting the system's ability to precisely target and puncture rat tail veins. Specifically, Figure \ref{fig:Ultrasound_Results}(a) illustrates the venipuncture procedure schematically, while Figure \ref{fig:Ultrasound_Results}(c) displays pre-puncture and post-puncture ultrasound images alongside the blood return status for each trial, further demonstrating the system's effectiveness.

The results of the \textit{in vivo} study underscore the exceptional precision and accuracy of the system in targeting and puncturing rat tail veins. These findings are particularly noteworthy given the inherent challenges associated with peripheral venous access in pediatric patients.

\section{Conclusion and Future Work}


This work introduces a newly designed robotic system, providing a novel case study for the design of robotic vascular access procedures. Specifically, this study aimed to enhance the accuracy of challenging vascular access procedures. Consequently, the focus of the experiments was rigorously testing the effectiveness and success rate of puncture procedures facilitated by ultrasound guidance. The system achieved a remarkable 100\% success rate in phantom studies and a commendable 95\% success rate \textit{in vivo} with Sprague Dawley rats. These outcomes significantly surpass those of conventional manual procedures, which typically yield success rates ranging from 70\% to 90\%. We ascribe this success to a superior mechanical integration design, integrated with the successful application of ultrasound technology.

Looking ahead, we aim to incorporate advanced computer vision techniques for real-time estimation of vein diameter. Additionally, we will prioritize the development of an intuitive user interface leveraging augmented and virtual reality technologies to enhance visualization and control. Although our current work is confined to phantom and animal studies, we plan to embark on large-scale clinical trials in the future to validate the RVA system and its associated workflows. With continuous improvement, this robotic system holds immense potential as an assistive tool in clinical practice, offering a safer and more efficient alternative to traditional manual intravenous vascular access methods. It has the capacity to improve patient outcomes and mitigate healthcare costs.

\bibliographystyle{ieeetr}
\balance
\bibliography{reference}

\end{document}